\documentclass[letterpaper, 10 pt, conference]{ieeeconf}  

\IEEEoverridecommandlockouts                              

\usepackage{graphicx}
\usepackage{amsmath}
\usepackage{amssymb}
\usepackage{listings}
\usepackage{algorithm}
\usepackage{algpseudocode}
\usepackage[table,xcdraw]{xcolor}
\usepackage{booktabs} 
\usepackage{caption}
\usepackage{subcaption}
\usepackage[colorlinks,
citecolor=blue 
]{hyperref}

\DeclareRobustCommand{\spatialone}{%
  \begingroup\normalfont
  \includegraphics[height=\fontcharht\font`\B]{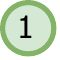}%
  \endgroup
}
\DeclareRobustCommand{\spatialtwo}{%
  \begingroup\normalfont
  \includegraphics[height=\fontcharht\font`\B]{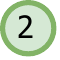}%
  \endgroup
}
\DeclareRobustCommand{\spatialthree}{%
  \begingroup\normalfont
  \includegraphics[height=\fontcharht\font`\B]{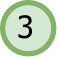}%
  \endgroup
}
\DeclareRobustCommand{\spatialfour}{%
  \begingroup\normalfont
  \includegraphics[height=\fontcharht\font`\B]{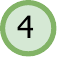}%
  \endgroup
}
\DeclareRobustCommand{\nonspatialfive}{%
  \begingroup\normalfont
  \includegraphics[height=\fontcharht\font`\B]{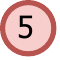}%
  \endgroup
}
\DeclareRobustCommand{\nonspatialsix}{%
  \begingroup\normalfont
  \includegraphics[height=\fontcharht\font`\B]{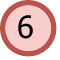}%
  \endgroup
}

\title{\LARGE \bf
Multi-Agent Path Finding with Prioritized Communication Learning
}

\author{Wenhao Li$^{\dag}$, Hongjun Chen$^{\dag}$, Bo Jin, Wenzhe Tan, Hongyuan Zha and Xiangfeng Wang$^{\ddag}$%
\thanks{${\dag}$ W. Li and H. Chen contribute equally. $^{\ddag}$ Corresponding author.}
\thanks{W. Li, H. Chen, B. Jin and X. Wang are with the School of Computer Science and Technology, East China Normal University, Shanghai 200062, China. X. Wang is also with Key Laboratory of Artificial Intelligence, Ministry of Education, Shanghai 200240, China.
W. Tan is with Geekplus Technology Co., Ltd., Beijing, 100101, China.
H. Zha is with School of Data Science and AIRS, The Chinese University of Hong Kong, Shenzhen 518172, China. 
}%
}

\begin{document}

\maketitle
\thispagestyle{empty}
\pagestyle{empty}

\begin{abstract}
Multi-agent pathfinding (MAPF) has been widely used to solve large-scale real-world problems, e.g., automation warehouses.
The learning-based, fully decentralized framework has been introduced to alleviate real-time problems and simultaneously pursue optimal planning policy. 
However, existing methods might generate significantly more vertex conflicts (or collisions), which lead to a low success rate or more makespan.
In this paper, we propose a PrIoritized COmmunication learning method (PICO), which incorporates the \textit{implicit} planning priorities into the communication topology within the decentralized multi-agent reinforcement learning framework.
Assembling with the classic coupled planners, the implicit priority learning module can be utilized to form the dynamic communication topology, which also builds an effective collision-avoiding mechanism.
PICO performs significantly better in large-scale MAPF tasks in success rates and collision rates than state-of-the-art learning-based planners.
\end{abstract}


\section{INTRODUCTION}\label{intro}

With the rapid development of low-cost sensors and computing devices, more and more manufacturing application scenarios can support the concurrent control of large-scale automated guided vehicles (AGVs)~\cite{Rubenstein2014ProgrammableSI,Howard2006ExperimentsWA}.
To achieve efficient large-scale transportation via AGVs, many efforts have been devoted to multi-agent path finding, which aims to plan paths for all agents~\cite{Silver2005CooperativeP,Berg2009ReciprocalNC,Sharon2012ConflictbasedSF,Felner2018AddingHT,Sartoretti2019PRIMALPV,Zhang2020LearningTC,Damani2021PRIMAL_2PV}.
A critical constraint is that the agents must follow their paths concurrently without collisions~\cite{Stern2019MultiAgentPD}.
Although classic ``optimal" planners~\cite{Silver2005CooperativeP,Berg2009ReciprocalNC,Sharon2012ConflictbasedSF,Felner2018AddingHT} 
can guarantee the completeness of the solution and collision-free under certain assumptions,
these planners have to re-plan for new scenarios, which cannot adequately satisfy the real-time requirements of realistic tasks.

\begin{figure}[h]
    \centering
    \begin{subfigure}[b]{0.22\textwidth}
        \includegraphics[width=\textwidth]{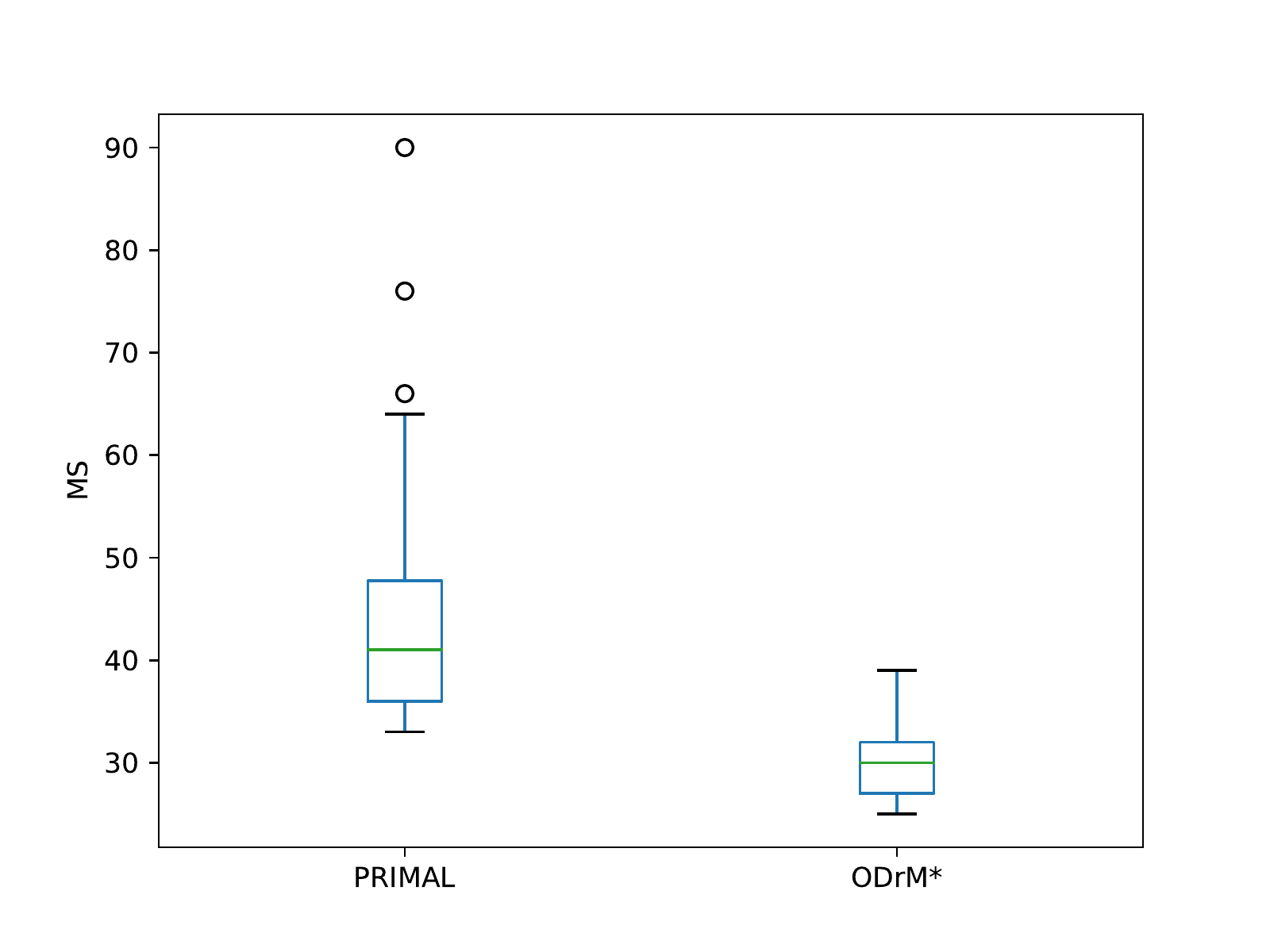}
        \caption{Makespans.}
    \end{subfigure}
    \begin{subfigure}[b]{0.21\textwidth}
        \centering
        \includegraphics[width=\textwidth]{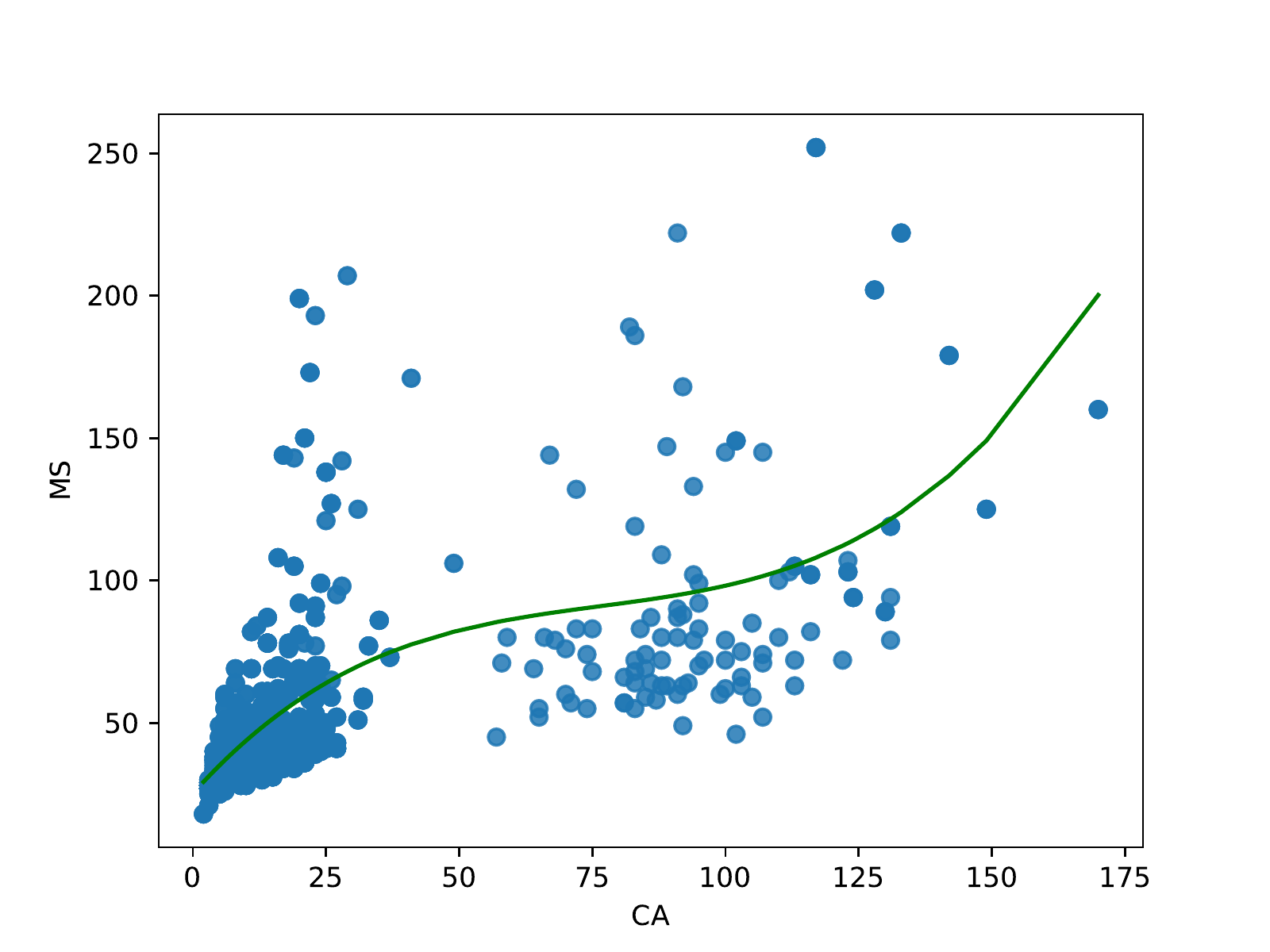}
        \caption{Collisions vs. Makespans.}
    \end{subfigure}
    \caption{(a)  Makespans (the maximum timesteps for all agents to reach the goals) of PRIMAL and ODrM*; (b) Each point represents the total number of collisions between each agent and other agents, and the makespan in the planned path obtained by PRIMAL.}
    \label{fig:collisions}
    \vspace{-20pt}
\end{figure}


Recently, many learning-based methods~\cite{Sartoretti2019PRIMALPV,Zhang2020LearningTC,Damani2021PRIMAL_2PV} were proposed to solve the above issues. 
Unlike classic planners that calculate a complete path for each agent based on global information, learning-based methods will learn a planning policy for each agent, which only employs local observations to decide one-step or limited length of paths.
These learning-based methods generally model the MAPF problem as a multi-agent reinforcement learning (MARL) problem.
The collision-free path planning can further be approximated in two ways.
The first is to introduce post-processing techniques to avoid collisions by adjusting each agent's individual paths~\cite{Zhang2020LearningTC}, however this scheme might be extremely time-consuming.
In a more ``soft'' manner, the second way is penalizing collisions by hand-crafted reward shaping \cite{Sartoretti2019PRIMALPV,Damani2021PRIMAL_2PV}, which do not assure that the solutions are collision-free.
But, its apparent shortcoming is that collision-free planning cannot usually be achieved.
From the experimental perspective, the agents following the obtained planning paths will frequently collide with each other, thereby increasing the length of the planned paths than classic planners due to their decentralized framework, e.g., PRIMAL~\cite{Sartoretti2019PRIMALPV} and ODrM*~\cite{Ferner2013ODrMOM} in Figure~\ref{fig:collisions}.
More importantly, each agent make independent decisions based on its local observations without any global information, which could lead to a deadlock state~\cite{Sartoretti2018DistributedRL}.

The key reason, that existing learning-based MAPF algorithms cannot achieve better collision avoidance and real-time simultaneously, lies in their fully decentralized framework.
For each agent, classic planners 
usually employ the full or partial information of others to achieve collision-free.
Therefore, this paper utilizes the decentralized path planning and centralized collision avoiding framework, and introduces the communication-based MARL to design our learning-based method.
However, one challenge is that a collision avoiding driven communication protocol becomes more challenging to obtain through this end-to-end communication learning scheme than classic planners~\cite{Lowe2019OnTP}.
Prioritized MAPF algorithms~\cite{Silver2005CooperativeP,Ma2019SearchingWC,Sturtevant2006ImprovingCP}, which usually assign a total priority to each agent and plan a path for each agent from high-priority to low-priority, are among the most efficient state-of-the-art MAPF methods for collision avoidance~\cite{Sharon2012ConflictbasedSF,Velagapudi2010DecentralizedPP,Wang2011MAPPAS}.
However, they determine a predefined total priority ordering of the agents, and might lead to lousy quality solutions or even fail to find any solutions~\cite{Ma2019SearchingWC}.
This motivates us to introduce the priority learning skill into communication learning scheme to learn a novel communication protocol with the collision alleviating ability.

\begin{figure*}
    \centering
    \includegraphics[width=0.96\textwidth]{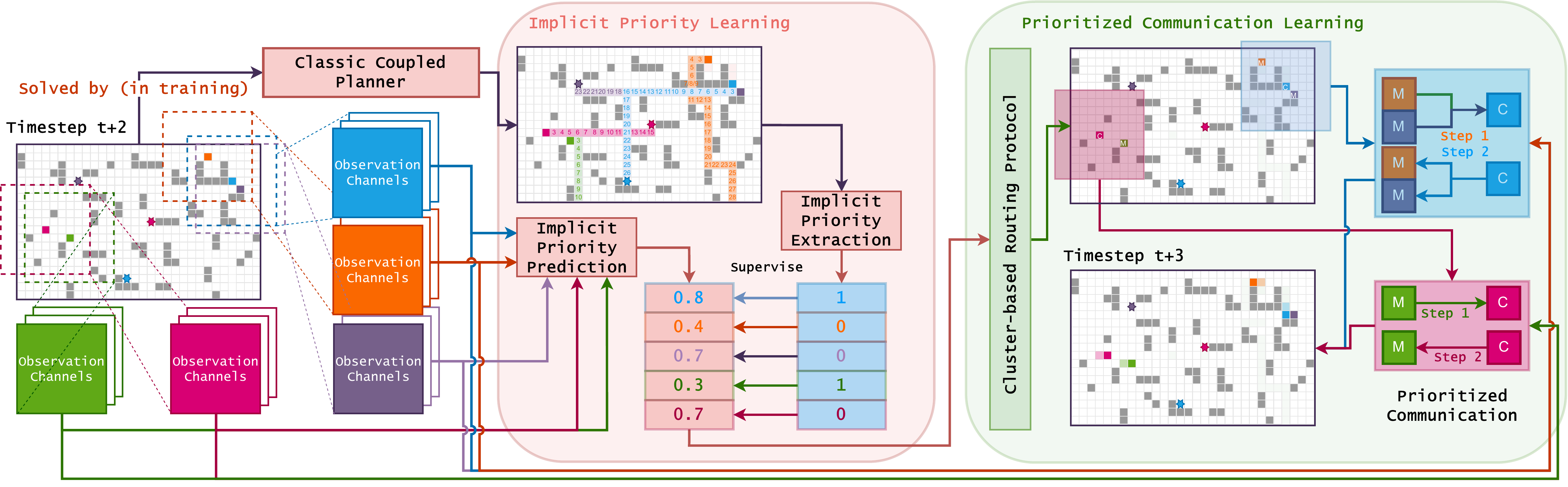}
    \caption{The overview of the PICO algorithm. PICO includes two sub-modules, i.e., \textit{implicit priority learning} and \textit{prioritized communication learning}. The ``C'' and ``M'' on the agent represent the central agent (high-priority) and member agents. The light-colored square in the map represents the agent's position at the last timestep.}
    \label{fig:overview}
    \vspace{-20pt}
\end{figure*}

In this paper, we propose the \textbf{P}r\textbf{I}oritized \textbf{CO}mmunication Learning method (\textbf{PICO}), to incorporate the \textit{implicit} planning priority into the communication topology within the communication-based MARL framework.
Specifically, PICO includes two phases, i.e., \textit{implicit priority learning} phase and \textit{prioritized communication learning} phase as shown in Figure~\ref{fig:overview}.
In the \textit{implicit priority learning}, PICO aims to build an auxiliary imitation learning task to predict the local priority of each agent by imitating classical coupled planner (e.g., ODrM*).
The information diffusion path will be determined by these implicit priorities, while the obtained asymmetric information and different implicit constraints will drive the agents to plan the nearly collision-free paths.
Besides benefiting to collision avoiding, the priorities can further be utilized to establish the collision avoiding driven communication protocol.
In the \textit{prioritized communication learning}, the obtained local priorities is used to generate a time-varying communication topology consists of agent clusters, where these priorities are considered as the weights of an ad-hoc routing protocol.
Each agent can learn its collision-reducing policy via the received messages, which are considered as the environment's perception and are encoded by the collision-avoiding communication topology.

The main contributions are summarized as follows:

\noindent 1) The decentralized path planning and centralized collision avoiding scheme is utilized via the communication-based MARL framework, to avoid collision more efficiently;

\noindent 2) The auxiliary imitation learning task is introduced to conduct implicit priority learning based on classic coupled planners' local priorities;

\noindent 3) The structured communication protocol with the collision alleviating ability is constructed by integrating the learnt priorities into communication learning scheme;

\noindent 4) Our proposed PICO performs significantly better in large-scale multi-agent path finding tasks in both success rate and collision rate than state-of-the-art learning-based planners.

\section{RELATED WORK}\label{sec: related-work}

\subsection{Learning-based Multi-Agent Path Finding}\label{sec: mapf}

Many works employed MARL or deep learning (DL) methods to solve the MAPF problem.
The typical PRIMAL method~\cite{Sartoretti2019PRIMALPV} introduced MARL firstly, and combined behavior cloning from a dynamically coupled planner to accelerate the training procedure. 
Further the PRIMALc~\cite{Zhiyao2020DeepRL} extended PRIMAL from 2D to 3D search space.
PRIMALc introduced ``agent modeling" technique to assist path planning by predicting the actions of other agents in a decoupled manner.
Recently, PRIMAL$_2$~\cite{Damani2021PRIMAL_2PV} extended PRIMAL to the lifelong MAPF (LMAPF) scenario, 
in which each agent will be immediately assigned a new goal once upon reaching their current goal.
\cite{Zhang2020LearningTC} proposed the MATS method which employed the multi-step ahead tree-search strategy in single-agent reinforcement learning (SARL) and imitation learning scheme to fit the results of the tree search strategy to solve the MAPF problem.
BitString~\cite{freed2020simultaneous} and DHC~\cite{Ma2021DistributedHM} introduced the communication-based learning scheme based on end-to-end training procedure, but under fixed communication topology.

\subsection{Prioritized Multi-Agent Path Finding}\label{sec: pmapf}
Prioritized planning~\cite{Erdmann1987MultipleMoving} can be considered as a decoupled approach, where agents are ordered by predefined priorities.
Fixed predefined priorities are assigned to all agents globally while all collisions are resolved according to these priors before the movements begin.
Specially, the priorities can be assigned arbitrarily~\cite{Silver2005CooperativeP,Warren1990MultipleRP,Bennewitz2002FindingAO} or handcrafted
~\cite{Erdmann1987MultipleMoving}.
Heuristic methods can also be utilized to assign priorities,
e.g., 
\cite{Berg2005PrioritizedMP,Buckley1989FastMP,Ferrari1998MultirobotMC}.
Further, some local orderings can assign temporary priorities to some agents in order to resolve collisions on the fly.
This kind of method requires all agents to follow their assigned paths, while if an impasse is reached the priorities are assigned dynamically to determine which agent to wait~\cite{ODonnell1989DeadlockfreeAC,Azarm1997ConflictfreeMO}.
In order to improve the solution quality, many existing works attempt to elaborately explore the space of all total priority orderings.
Some works explored this space randomly by generating several total priorities ordering as part of a hill-climbing scheme~\cite{Bennewitz2002FindingAO}, instead of ergodically which can be enumerated for only up to
$3$ agents~\cite{Azarm1997ConflictfreeMO}.
Recently, CBSw/P~\cite{Ma2019SearchingWC} enumerated the sub-space (of all local priority orderings) as part of a conflict-driven combinatorial search framework.

\section{PRELIMINARIES}\label{sec:pre}

\noindent{\em{Multi-Agent Path Finding}}.
Classical MAPF can be formalized as follows.
The input to a MAPF problem with $N$ agents is a tuple $\langle G, s, g\rangle$ where $G=(V, E)$ denotes an undirected graph, and $s,g:[1, \ldots, N] \rightarrow V$ maps an agent to a source and goal vertex.
Each agent can either move to an adjacent vertex or wait at its current vertex.
A sequence of actions $\tau$ is a {\em{single-agent plan}} for agent $i$ iff executing this sequence of actions in $s(i)$ results in being at $g(i)$, that is, iff $\tau^i[|\tau|]=g(i)$.
The {\em{solution}} is a set of $N$ single-agent plans, one for each agent.
The overarching goal of MAPF solvers is to find a solution that can be executed without collisions.
Let $(\tau^i,\tau^j)$ be a pair of single-agent plans, a \textit{vertex conflict} between $\tau^i$ and $\tau^j$ occurs iff according to their plans, the agents are planned to occupy the same vertex at the same timestep.
This paper uses one of the most common functions, \textit{makespan} which is defined as $\max_{1 \leq i \leq N}\left|\tau^i\right|$, for evaluating a solution as in classical MAPF~\cite{Stern2019MultiAgentPD}.

\noindent{\em{Partially Observable Stochastic Game}}.
POSG~\cite{hansen2004dynamic} can be denoted as a tuple $
\langle \mathcal{X}, \mathcal{S}, \left\{ \mathcal{A}^i \right\}_{i=1}^{n}, \left\{ \mathcal{O}^i \right\}_{i=1}^{n}, \mathcal{P}, \mathcal{E}, \left\{ \mathcal{R}^i \right\}_{i=1}^{n} \rangle,
$ where $n$ is the number of agents, $\mathcal{X}$ is the agent space, $\mathcal{S}$ is the state space,
$\mathcal{A}^i$ is the action space of agent $i$, $\boldsymbol{\mathcal{A}}=\mathcal{A}^1\times\mathcal{A}^2\times\cdots\times\mathcal{A}^n$ is the joint action space,
$\mathcal{P}(s^{\prime}|s, \boldsymbol{a})$ is the state transition probability function,
$\mathcal{O}^i$ is the observation space of agent $i$, $\boldsymbol{\mathcal{O}}=\mathcal{O}^1\times\mathcal{O}^2\times\cdots\times\mathcal{O}^n$ is the joint observation space, $\mathcal{E}(\boldsymbol{o}|s)$ is the observation emission probability function,
and $\mathcal{R}^i: \mathcal{S}\times\boldsymbol{\mathcal{A}}\times\mathcal{S} \rightarrow {\mathbb{R}}$ is the reward function of agent $i$.
The objective of each agent is to maximize its expected total return during the game.

\section{THE PROPOSED PICO METHOD}\label{sec:method}
In this section, we propose our prioritized communication learning method, to incorporate the \textit{implicit} planning priority into the communication topology learning procedure within the communication-based MARL framework.
As shown in Figure~\ref{fig:overview}, PICO includes two phases, i.e., the \textit{implicit priority learning} and the \textit{prioritized communication learning}.
These two phases are executed randomly during training procedure.

\subsection{Implicit Priority Learning}

The implicit priority learning phase aims to imitatively learn implicit priorities from classical coupled planner.
In this paper, we utilize the ODrM* as the targeted classical coupled planner, which is considered as the state-of-the-art MAPF method~\cite{Ferner2013ODrMOM}.
Further, we introduce the behavior cloning~\cite{Sartoretti2019PRIMALPV,Zhang2020LearningTC,Damani2021PRIMAL_2PV,Zhiyao2020DeepRL} to enhance the training procedure efficiency from the perspectives of both convergence speed and training performance.
To emphasis, the behavior cloning shares the same low-level parameters with the implicit priority learning model, while this mutual restriction will benefit to the effectiveness of both implicit priority learning and behavior cloning~\cite{Levine2014GuidedPS,Levine2014LearningNN}.
As follows, we will introduce the detailed procedure of this phase, while training datasets need to be constructed first.

{\em{Imitation Dataset Construction}}:
Each agent's observation and action spaces are the similar as previous work~\cite{Sartoretti2019PRIMALPV,Damani2021PRIMAL_2PV}.
The training sets $\mathcal{D}_{imp}$, $\mathcal{D}_{imt}$ for the implicit priority learning and the behavior cloning training respectively can be constructed based on the solutions of the classical coupled planner, i.e., the ODrM*.
In every priority learning episode, given the $k$-th randomly generated MAPF instance $\mathcal{F}_k:=\langle G_k, s_k, g_k\rangle$, the ODrM* method will obtain a solution, i.e., the set of all single-agent plans $\{\tau^i_k\}$($i=1,\cdots,N$).
By assuming the makespan of the solution is $T_k$, all position-position pairs $\{\tau^i_{k}[t], \pi^i_t(\tau^i_{k}[t])\}$($i=1,\cdots,N$) can be calculated through unfolding all single-agent plans $\{\tau^i_k\}$.
At each timestep $t$, current positions $\{\tau^i_{k}[t]\}$ and corresponding goals $g_k$ of all agents can be combined to construct $\mathcal{D}_{imt}$, i.e., $$\mathcal{D}_{imt}:=\{o_m, a_m\}_{m=1}^{|\mathcal{D}_{imt}|},\quad {\textstyle{|\mathcal{D}_{imt}|=N\times\sum_{k=1}^K T_k}},$$where $o_m$ denotes the observations constructed by $\{\tau^i_k\}$($i=1,\cdots,N$) within the same scheme as~\cite{Sartoretti2019PRIMALPV,Damani2021PRIMAL_2PV}, and $a_m$ denotes the actions obtained by calculating the difference between two consecutive positions $\{\tau^i_{k}[t], \pi^i_t(\tau^i_{k}[t])\}$.

Further, we construct the ground-truth implicit priorities of each observation $o_m$ according to the following steps.
Based on the observations of each agent $i$ with its goal $g_k(i)$, the next optimal action set $\mathcal{A}^i_{k}[t+1]$ is established through a greedy strategy (the next optimal action is the action that can touch the goal at the fastest speed without considering other agents).
If the agent is at the goal, then the optimal action set is empty.
Considering the optimal action set and the output action of ODrM*, the implicit priority $p_m$ of $o_m$ can be calculated via following two rules: 
i) the action is ``no-movement'': the priority $p_m$ is set to be high ($p_m=1$) if the optimal action set is empty, otherwise is set to be low ($p_m=0$);
ii) the action is ``movement'': the priority $p_m$ is set to be high ($p_m=1$) if the action is in the optimal action set, otherwise is low ($p_m=0$).
The dataset $\mathcal{D}_{imt}$ is constructed accordingly, $$\mathcal{D}_{imp}:=\{o_m, p_m\}_{m=1}^{|\mathcal{D}_{imp}|},\quad {\textstyle{|\mathcal{D}_{imp}|=N\times\sum_{k=1}^K T_k}}.$$

\begin{figure}
    \centering
    \includegraphics[width=0.48\textwidth]{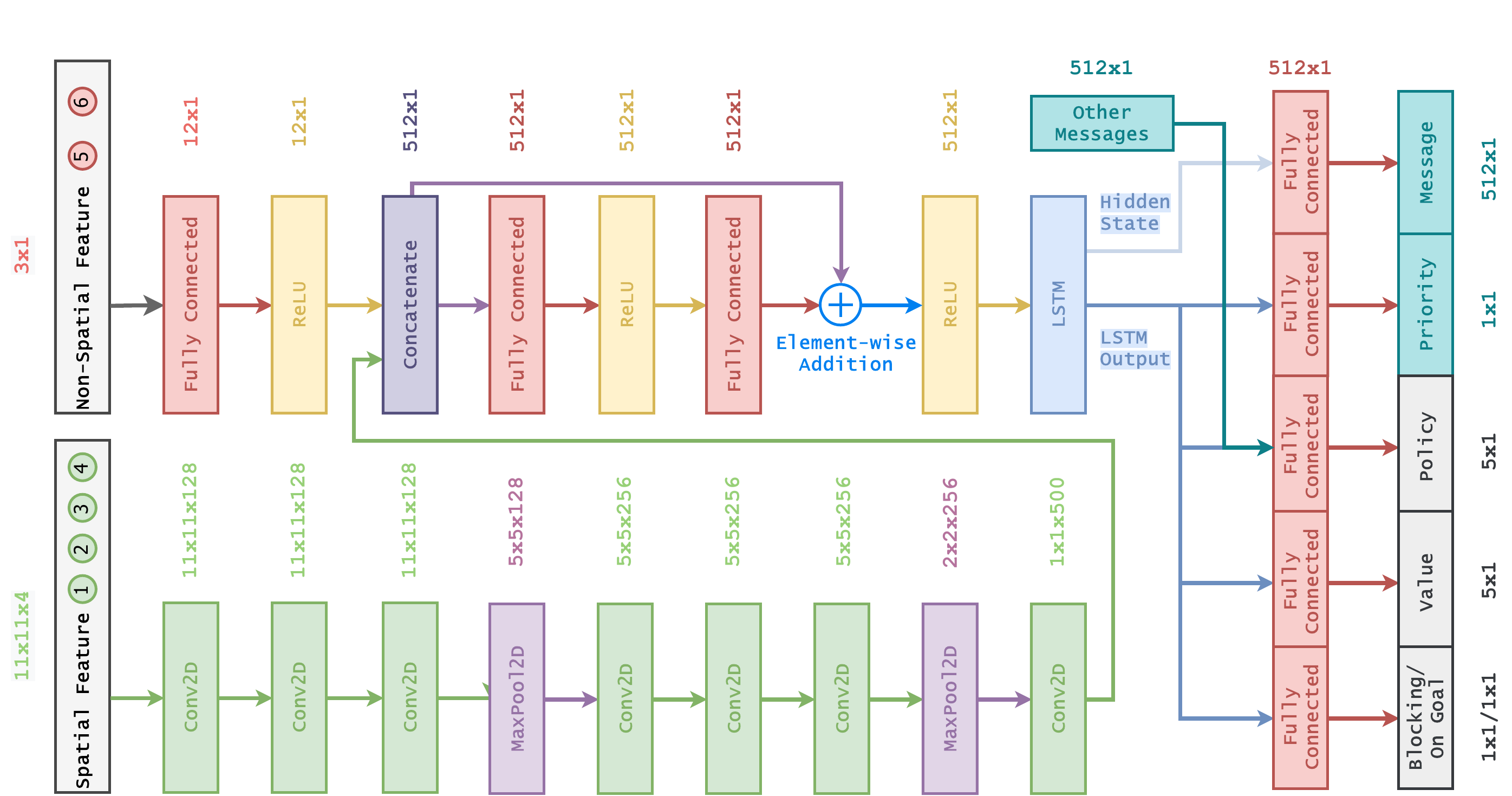}
    \caption{The network structure of priority prediction model and behavior cloning model. The local information of each agent is broken down into channels (\spatialone{}\spatialtwo{}\spatialthree{}\spatialfour{}). Each agent also has access to non-spatial features (\nonspatialfive{}\nonspatialsix{}).}
    \label{fig:net}
    \vspace{-20pt}
\end{figure}

{\em{Network Structure and Training Procedure}}:
With the obtained training dataset, we further establish the implicit priority learning model and the behavior cloning model structures as in Figure~\ref{fig:net}.
These two neural networks share the same low-layer parameters with different output-heads.
It is worth noting that messages from other agents are set to be the input of the neural network in Figure~\ref{fig:net}.
The corresponding messages are generated through the outputs (the ``message'' output-head in the Figure~\ref{fig:net}) of other agents.
The objective agents of sending the obtained messages are 
determined by the dynamic communication topology generated in the following prioritized communication learning phase.

The above-mentioned neural network is parameterized by $\theta$ and denoted as $\mathcal{I}_{\theta}$.
The output message, predicted priority and prediction action are denoted as $e_m:=\mathcal{I}^e_{\theta}(o_m)$, $\hat{p}_m:=\mathcal{I}^p_{\theta}(o_m)$ and $\hat{a}_m:=\mathcal{I}^{\pi}_{\theta}(o_m, \{e_m\})$ respectively.
In the training procedure\footnote{The predicted priorities are utilized to construct the communication topology in the prioritized communication learning phase. To strengthen the stability of the communication learning procedure, PICO introduces a similar target network as in~\cite{Mnih2015HumanlevelCT} and~\cite{Grill2020BootstrapYO}. 
The target network is updated with a slow-moving average of the online network.}, the loss function is set to be
\begin{equation}\label{eq:ploss}
    \mathcal{L}_{p} = \alpha_{imp}\mathcal{L}_{imp} + \mathcal{L}_{imt},
\end{equation}
where $\alpha_{imp}$ denotes the penalty parameter and
$$
\begin{aligned}
        \mathcal{L}_{imp} \!\!&=\!\! {\textstyle{-\frac{1}{|\mathcal{D}_{imp}|} \!\!\sum_{m=0}^{|\mathcal{D}_{imp}|-1}\! \left[ p_m \log(\hat{p}_m) + (1\!-\!p_m)\log(1\!-\!\hat{p}_m)\right],}} \\
        \mathcal{L}_{imt} \!\!&=\!\! {\textstyle{-\frac{1}{|\mathcal{D}_{imt}|} \sum_{m=0}^{|\mathcal{D}_{imt}|-1} \sum_{c=0}^{|\mathcal{A}|-1} a_m[c]\log(\hat{a}_m[c]),}}
\end{aligned}$$where $|\mathcal{A}|$, $a_m[c]$ and $\hat{a}_m[c]$ represent the action space's size, the $c$-th element in $a_m$ and $\hat{a}_m$ respectively.

For the multi-agent learning problem, one of the key challenges is to encourage each agent to be selfless, while 
it might be detrimental to the immediate return maximization. 
This challenge is usually denoted as social dilemma~\cite{shoham2008multiagent} and exists conspicuously in MAPF~\cite{Sartoretti2018DistributedRL}.
Many algorithms have been proposed to alleviate this problem in multi-agent learning~\cite{Peysakhovich2018ConsequentialistCC,Peysakhovich2018ProsocialLA,Hughes2018InequityAI,Jaques2019SocialIA}, however rarely few works pay attention to 
MAPF problem.
PICO method introduces additional {\em{blocking prediction}} tasks together with message exchanging 
to alleviate the social dilemma similar with the learning-based PRIMAL~\cite{Sartoretti2019PRIMALPV}.
Furthermore, an {\em{on-goal prediction}} task is introduced to establish an accurate perception of whether the agent has reached the goal.
Ablation studies show that the blocking prediction task can obtain better results when paired with the on-goal prediction task.
These two tasks can
provide more supervision signals from different perspectives
for both implicit priority learning and behavior cloning learning, which also can help to learn better representations of the local observations.

\subsection{Prioritized Communication Learning}



{\em{Dynamic Prioritized Communication Topology}}:
In order to achieve proper integration of implicit priority and communication learning, PICO introduces the classic ad-hoc routing protocol, i.e., \textit{cluster-based routing protocol} (CBRP~\cite{cbrp}) into the overall framework.
CBRP employs the ``weight'' attributes attached to each agent to construct a cluster-based communication topology, which can be dynamically updated once the agent weights change.
The high-weight agent becomes the center of the cluster, receives messages from all low-weight agents in the cluster, and broadcasts it to all low-weight agents after post-processing. 
Agents with low weights become affiliated members and make decisions based on the messages broadcast from the cluster‘s central agent. 
PICO chooses CBRP as the fundamental communication topology construction algorithm in MAPF because as long as the artificially defined weights in CBRP are replaced by the obtained implicit priorities,
the integration of priority and communication learning can be properly realized.

\begin{figure*}
    \centering
    \includegraphics[width=0.985\textwidth,angle=0]{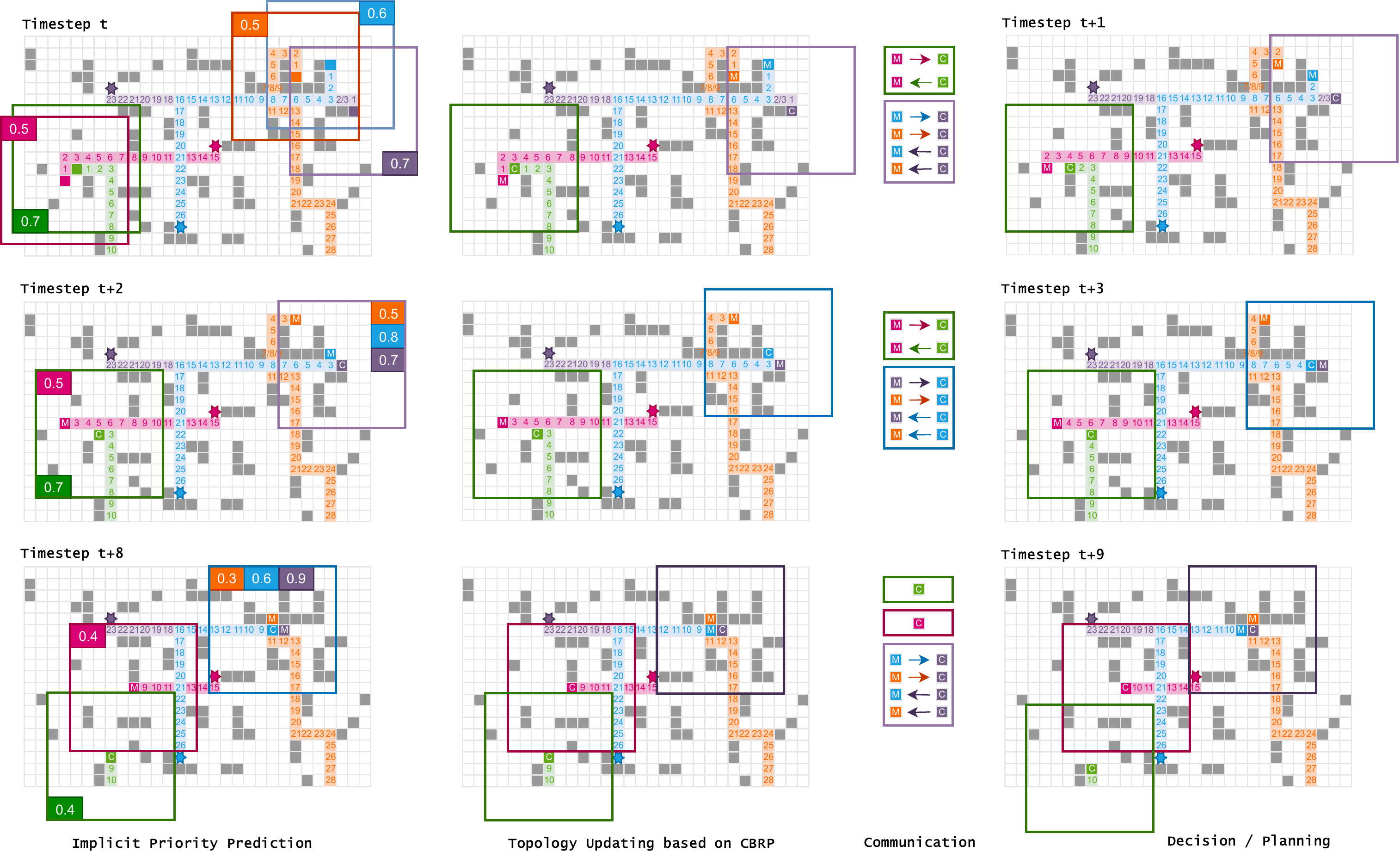}
    \caption{Schematic diagram of the process from priority prediction to decision in the prioritized communication learning phase at different timesteps. Each round in each row consists of four steps. First, PICO predicts each agent's local priority via priority learning model (Col. 1). Then, agents are clusterd by a priority-based routing protocol, CBRP (Col. 2). All clusters, which are consist of one central agent (mark as `C') and several member agents (mark as 'M'), form the dynamic topology together. After that, all agents conduct collision avoiding driven communication through the topology (Col. 3), and finally make their one-step planning (Col. 4).}
    \label{fig:cbrp}
    \vspace{-20pt}
\end{figure*}

Specifically, CBRP usually takes a cluster radius $d$ to establish the structured communication topology.
Each member agent can confirm whether others have larger weights or contain central agents within its receptive area.
If no such agent is found, this agent is elected as a central agent; 
otherwise, its role is kept. 
Meanwhile, each central agent checks whether other centrals exist in radius $d$. 
If no such agent is found or the found centrals' weights are smaller than its weights, 
its role is kept; 
otherwise, it downgrades to a member agent. 
After a sufficient number of rounds, 
all agents are separated into clusters with one central agent in each cluster.
This cluster-based communication topology inherits the classic planners' idea of avoiding planning only in the agent's neighborhood but also achieves the centralization-decentralization trade-off in MARL.
The illustration diagram of CBRP is shown in Figure~\ref{fig:cbrp}.

{\em{Network Structure and Training Procedure}}: PICO employs the A3C~\cite{Mnih2016AsynchronousMF} as the backbone of communication-based MARL.
The observation space is the same as the implicit priority learning phase.
We employ the different action space from the prioritized communication learning phase for the reason that
the agent may execute invalid actions.
If there are static obstacles, other agents, or beyond the map's boundary at the position reached by an agent's following action, such an action is called invalid.
Actions are sampled only from valid actions in training.
Previous results in \cite{Vinyals2017StarCraftIA,Sartoretti2018DistributedRL} show that this technique enables a more stable training procedure, compared with giving negative rewards to agents for selecting invalid moves.
Moreover, the reward function is same as~\cite{Sartoretti2019PRIMALPV}.
The network structure is consistent with previous phase as shown in Figure~\ref{fig:net}.
In the training procedure, the parameters of ``priority'', ``blocking'', and ``on-goal'' three output-heads are fixed.
The predicted value by the ``value'' output-head is denoted as $\hat{v}_m:=\mathcal{I}^v_{\theta}(o_m, a_m)$.
The parameters are updated to minimize the Bellman error between the predicted value and the discounted return, i.e., 
$$
\mathcal{L}_{v} = {\textstyle{\frac{1}{|B|} \sum_{m=0}^{|B|-1} \left( \hat{v}_m - R_m \right)^2}},\quad R_m:={\textstyle{\sum_{i=0}^{k} \gamma^i r_{t+i},}}
$$
where
$t$ denotes the timestep of $o_m$, $|B|$ is the mini-batch size, and $\gamma$ is the discount factor.
An approximation of the advantage function $A_\theta(o_m, a_m)$ by bootstrapping the value function is used to update the policy.
An entropy term $\mathcal{H}(\mathcal{I}^{\pi}_\theta(o_m,\{e_m\}))$ is further introduced to encourage exploration and discourage premature convergence~\cite{Haarnoja2018SoftAO}. The loss function of this prioritized communication learning phase can be denoted as
$$
    \begin{aligned}
    \mathcal{L}_{\pi}= - {\textstyle{\frac{1}{|B|} \sum_{m=0}^{|B|-1}}} &\big[ \log \mathcal{I}^{\pi}_\theta \big(a_{m} \mid o_m,\{e_m\} \big) A_\theta(o_m, a_m) \\
    &\quad -\sigma_{\mathcal{H}} \cdot \mathcal{H} \big(\mathcal{I}^{\pi}_\theta(o_m,\{e_m\}) \big) \big],
    \end{aligned}
$$
where $A_\theta(o_m, a_m)=\sum_{i=0}^{k-1} \gamma^{i} r_{t+i}+\gamma^{k} \mathcal{I}^v_\theta\left(o_{k+t}\right)-\hat{v}_m$ and $\sigma_{\mathcal{H}}$ denotes a small positive entropy weight.

        
    

\section{NUMERICAL EXPERIMENTS}\label{sec:exp}
In this section, we conduct the numerical experiments based on the modified gridworld environment of the Asprilo benchmark~\cite{Nguyen2017GeneralizedTA}, which is commonly adopted by classic or learning-based MAPF solvers. First, some fundamental settings are discussed in the following.


\noindent{\em{Environments}}. 
In training stage, the sizes of the gridworld are randomly selected as $10$ (twice as likely), $40$, or $70$, when each episode starts.
The number of agents is fixed at $8$.
The obstacle density is randomly selected from a triangular distribution between $0\%$ and $50\%$.
In testing stage, the gridworld size is fixed at $20$, the number of agents varies among $8$, $16$, $32$, and $64$.
The obstacle density is selected from $0\%$ to $30\%$ with an equal interval $10\%$.

\begin{table}[]
\centering
\caption{Evaluation measurements.}
\label{tab:measurements}
\resizebox{\linewidth}{!}{%
\begin{tabular}{|c|c|}
\hline
\rowcolor[HTML]{9B9B9B} 
{\color[HTML]{000000} Measurement}     & {\color[HTML]{000000} Description}               \\ \hline
\textbf{Collision with agents (CA)} & Number of vertex conflicts with other agents. \\ \hline
\textbf{Collision with obstacles (CO)} & Number of collisions with obstacles.             \\ \hline
\textbf{Success rate (SR)}             & Number of successful solutions.                  \\ \hline
\textbf{Makespan (MS)}                 & As defined in Section 2.1                        \\ \hline
\textbf{Collision rate (CR)}           & CA / MS                                          \\ \hline
\textbf{Total moves (TM)}              & Number of non-still actions taken by all agents. \\ \hline
\end{tabular}%
}
\vspace{-10pt}
\end{table}

\noindent{\em{Measurements}}.
To evaluate the performance from various aspects, the average values of $100$ randomly generated cases/environments on $6$ measurements are shown in Table~\ref{tab:measurements}.


\noindent{\em{Baseline Methods and Training Details}}.
Baseline methods include classic planners and learning-based (or MARL-based) methods.
Based on the results shown in recent works~\cite{Sartoretti2019PRIMALPV,Damani2021PRIMAL_2PV},
we selects ODrM*
with inflation factor $\epsilon=2.0$ as the classic planner baseline.
In addition, we choose PRIMAL~\cite{Sartoretti2019PRIMALPV} and DHC~\cite{Ma2021DistributedHM} as learning-based baselines.
For ODrM*, a timeout of $12$ seconds is used to match previous results~\cite{Wagner2015SubdimensionalEF}.
For PRIMAL and DHC, the agents plan individual paths are for up to $256$ timesteps.
The training procedure is performed at the same device as baselines.
See Table~\ref{table:picohp} for more details.
\begin{table}[]
\centering
\caption{Settings of PICO used in experiments.}
\label{table:picohp}
\resizebox{\linewidth}{!}{%
\begin{tabular}{|c|c|c|c|c|c|c|c|}
\hline
\rowcolor[HTML]{9B9B9B} 
\textbf{Name} & \textbf{Value} & \textbf{Name} & \textbf{Value} & \cellcolor[HTML]{9B9B9B}\textbf{Name} & \cellcolor[HTML]{9B9B9B}\textbf{Value} & \cellcolor[HTML]{9B9B9B}\textbf{Name} & \cellcolor[HTML]{9B9B9B}\textbf{Value} \\ \hline
\textbf{parallel envs} & 12 & \textbf{episode length} & 256 & \textbf{lr decay} & 5e-5 & \textbf{policy reg} & L2 \\ \hline
\textbf{steps/update} & 128 & \textbf{batch size} & 128 & \textbf{policy reg coef} & 1e-3 & \textbf{value reg} & L2 \\ \hline
\textbf{batch handling} & shuffle & \textbf{value loss} & MSE & \textbf{value reg coef} & 1.0 & \textbf{value grad clip} & 10*num of agents \\ \hline
\textbf{imitation loss} & CrossEntropy & \textbf{discount} & 0.95 & \textbf{policy grad clip} & 0.5 & \textbf{temperture} & 0.5 \\ \hline
\textbf{optimizer} & Nadam & \textbf{lr} & 2e-4 & \textbf{warm up episodes} & 2500 & \textbf{CBRP neighbors} & 11 \\ \hline
\textbf{moment} & 0.9 & \textbf{epsilon} & 1e-7 & \textbf{CBRP waiting time} & 1 & \textbf{LSTM embedding} & 512 \\ \hline
\end{tabular}%
}
\vspace{-10pt}
\end{table}

\begin{table*}[ht!]
\centering
\caption{The comparison of different algorithms in the environment with $8,16,32,64$ agents, $20$-sized, and various obstacle densities. All results are the average values of $100$ randomly generated cases/environments.
In each column, the {\color{red}{red}} and {\color{green}{green}} numbers stand for the {\color{red}{best}} and {\color{green}{our}} results, $\uparrow$ stands that bigger is better, and $\downarrow$ stands that smaller is better.}
\label{tab:main-rst}
\resizebox{\textwidth}{!}{%
\begin{tabular}{|c|c|c|c|c|c|c|c|c|c|c|c|c|c|c|c|c|c|c|c|c|c|c|c|c|}
\hline
 &
  \multicolumn{24}{c|}{\textbf{8 Agents (0\%,10\%,20\%,30\% Obstacle Densities)}} \\ \cline{2-25} 
{\textbf{Methods}} &
  \multicolumn{4}{c|}{\textbf{CA $\downarrow$}} &
  \multicolumn{4}{c|}{\textbf{CO $\downarrow$}} &
  \multicolumn{4}{c|}{\textbf{SR $\uparrow$}} &
  \multicolumn{4}{c|}{\textbf{MS $\downarrow$}} &
  \multicolumn{4}{c|}{\textbf{CR $\downarrow$}} &
  \multicolumn{4}{c|}{\textbf{TM $\downarrow$}} \\ \hline
\textbf{ODrM*~\cite{Ferner2013ODrMOM}} &
  {\color[HTML]{CB0000} \textbf{0.0}} &
  {\color[HTML]{CB0000} \textbf{0.0}} &
  {\color[HTML]{CB0000} \textbf{0.0}} &
  {\color[HTML]{CB0000} \textbf{0.0}} &
  {\color[HTML]{CB0000} \textbf{0.0}} &
  {\color[HTML]{CB0000} \textbf{0.0}} &
  {\color[HTML]{CB0000} \textbf{0.0}} &
  {\color[HTML]{CB0000} \textbf{0.0}} &
  {\color[HTML]{CB0000} \textbf{100.0}} &
  {\color[HTML]{CB0000} \textbf{100.0}} &
  {\color[HTML]{CB0000} \textbf{100.0}} &
  {\color[HTML]{CB0000} \textbf{100.0}} &
  {\color[HTML]{CB0000} \textbf{25.55}} &
  {\color[HTML]{CB0000} \textbf{23.95}} &
  {\color[HTML]{CB0000} \textbf{25.86}} &
  {\color[HTML]{CB0000} \textbf{29.45}} &
  {\color[HTML]{CB0000} \textbf{0.0}} &
  {\color[HTML]{CB0000} \textbf{0.0}} &
  {\color[HTML]{CB0000} \textbf{0.0}} &
  {\color[HTML]{CB0000} \textbf{0.0}} &
  {\color[HTML]{CB0000} \textbf{112.75}} &
  {\color[HTML]{CB0000} \textbf{108.83}} &
  {\color[HTML]{CB0000} \textbf{117.49}} &
  {\color[HTML]{CB0000} \textbf{117.38}} \\ \hline
\textbf{PRIMAL~\cite{Sartoretti2019PRIMALPV}} &
  1.94 &
  3.02 &
  3.03 &
  5.98 &
  {\color[HTML]{CB0000} \textbf{0.0}} &
  0.01 &
  0.01 &
  {\color[HTML]{CB0000} \textbf{0.0}} &
  93.0 &
  90.0 &
  48.0 &
  15.0 &
  34.86 &
  62.69 &
  148.59 &
  234.22 &
  0.06 &
  0.05 &
  0.02 &
  0.03 &
  221.3 &
  223.01 &
  345.09 &
  565.11 \\ \hline
\textbf{DHC~\cite{Ma2021DistributedHM}} &
  1.66 &
  2.72 &
  3.74 &
  4.17 &
  {\color[HTML]{CB0000} \textbf{0.0}} &
  {\color[HTML]{CB0000} \textbf{0.0}} &
  {\color[HTML]{CB0000} \textbf{0.0}} &
  {\color[HTML]{CB0000} \textbf{0.0}} &
  91.0 &
  87.0 &
  {\color[HTML]{009901} \textbf{55.0}} &
  11.0 &
  33.7 &
  63.0 &
  136.89 &
  241.55 &
  0.05 &
  0.04 &
  0.03 &
  0.02 &
  271.56 &
  239.66 &
  401.48 &
  638.62 \\ \hline
\textbf{PICO-heuristic} &
  2.81 &
  2.7 &
  3.68 &
  3.4 &
  {\color[HTML]{CB0000} \textbf{0.0}} &
  {\color[HTML]{CB0000} \textbf{0.0}} &
  {\color[HTML]{CB0000} \textbf{0.0}} &
  {\color[HTML]{CB0000} \textbf{0.0}} &
  90.0 &
  86.0 &
  50.0 &
  10.0 &
  33.57 &
  59.81 &
  141.94 &
  236.99 &
  0.05 &
  0.04 &
  0.03 &
  0.02 &
  248.19 &
  252.88 &
  451.09 &
  650.62 \\ \hline
\textbf{PICO} &
  {\color[HTML]{009901} \textbf{0.59}} &
  {\color[HTML]{009901} \textbf{0.62}} &
  {\color[HTML]{009901} \textbf{1.31}} &
  {\color[HTML]{009901} \textbf{2.32}} &
  {\color[HTML]{CB0000} \textbf{0.0}} &
  {\color[HTML]{CB0000} \textbf{0.0}} &
  {\color[HTML]{CB0000} \textbf{0.0}} &
  {\color[HTML]{CB0000} \textbf{0.0}} &
  {\color[HTML]{CB0000} \textbf{100.0}} &
  {\color[HTML]{009901} \textbf{96.0}} &
  {\color[HTML]{009901} \textbf{55.0}} &
  {\color[HTML]{009901} \textbf{25.0}} &
  {\color[HTML]{009901} \textbf{27.45}} &
  {\color[HTML]{009901} \textbf{41.81}} &
  {\color[HTML]{009901} \textbf{134.7}} &
  {\color[HTML]{009901} \textbf{204.83}} &
  {\color[HTML]{009901} \textbf{0.02}} &
  {\color[HTML]{009901} \textbf{0.02}} &
  {\color[HTML]{009901} \textbf{0.01}} &
  {\color[HTML]{009901} \textbf{0.01}} &
  {\color[HTML]{009901} \textbf{123.96}} &
  {\color[HTML]{009901} \textbf{142.69}} &
  {\color[HTML]{009901} \textbf{290.13}} &
  {\color[HTML]{009901} \textbf{463.41}} \\ \hline
 &
  \multicolumn{24}{c|}{\textbf{16 Agents (0\%,10\%,20\%,30\% Obstacle Densities)}} \\ \cline{2-25} 
{\textbf{Methods}} &
  \multicolumn{4}{c|}{\textbf{CA $\downarrow$}} &
  \multicolumn{4}{c|}{\textbf{CO $\downarrow$}} &
  \multicolumn{4}{c|}{\textbf{SR $\uparrow$}} &
  \multicolumn{4}{c|}{\textbf{MS $\downarrow$}} &
  \multicolumn{4}{c|}{\textbf{CR $\downarrow$}} &
  \multicolumn{4}{c|}{\textbf{TM $\downarrow$}} \\ \hline
\textbf{ODrM*~\cite{Ferner2013ODrMOM}} &
  {\color[HTML]{CB0000} \textbf{0.0}} &
  {\color[HTML]{CB0000} \textbf{0.0}} &
  {\color[HTML]{CB0000} \textbf{0.0}} &
  {\color[HTML]{CB0000} \textbf{0.0}} &
  {\color[HTML]{CB0000} \textbf{0.0}} &
  {\color[HTML]{CB0000} \textbf{0.0}} &
  {\color[HTML]{CB0000} \textbf{0.0}} &
  {\color[HTML]{CB0000} \textbf{0.0}} &
  {\color[HTML]{CB0000} \textbf{100.0}} &
  {\color[HTML]{CB0000} \textbf{100.0}} &
  {\color[HTML]{CB0000} \textbf{100.0}} &
  {\color[HTML]{CB0000} \textbf{99.0}} &
  {\color[HTML]{CB0000} \textbf{26.54}} &
  {\color[HTML]{CB0000} \textbf{27.06}} &
  {\color[HTML]{CB0000} \textbf{28.51}} &
  {\color[HTML]{CB0000} \textbf{33.47}} &
  {\color[HTML]{CB0000} \textbf{0.0}} &
  {\color[HTML]{CB0000} \textbf{0.0}} &
  {\color[HTML]{CB0000} \textbf{0.0}} &
  {\color[HTML]{CB0000} \textbf{0.0}} &
  {\color[HTML]{CB0000} \textbf{216.35}} &
  {\color[HTML]{CB0000} \textbf{224.4}} &
  {\color[HTML]{CB0000} \textbf{228.96}} &
  {\color[HTML]{CB0000} \textbf{265.0}} \\ \hline
\textbf{PRIMAL~\cite{Sartoretti2019PRIMALPV}} &
  6.59 &
  8.29 &
  11.63 &
  17.64 &
  0.02 &
  0.04 &
  0.06 &
  0.06 &
  92.0 &
  88.0 &
  50.0 &
  3.0 &
  57.47 &
  71.82 &
  176.16 &
  249.37 &
  0.12 &
  0.11 &
  0.07 &
  0.07 &
  481.63 &
  510.11 &
  765.89 &
  1396.23 \\ \hline
\textbf{DHC~\cite{Ma2021DistributedHM}} &
  7.28 &
  8.95 &
  11.75 &
  18.7 &
  {\color[HTML]{CB0000} \textbf{0.0}} &
  0.06 &
  0.11 &
  0.13 &
  94.0 &
  88.0 &
  48.0 &
  5.0 &
  54.02 &
  71.19 &
  169.18 &
  242.06 &
  0.13 &
  0.12 &
  0.07 &
  0.08 &
  476.51 &
  477.0 &
  822.23 &
  1503.58 \\ \hline
\textbf{PICO-heuristic} &
  5.93 &
  9.71 &
  12.28 &
  17.54 &
  {\color[HTML]{CB0000} \textbf{0.0}} &
  0.04 &
  0.04 &
  0.28 &
  92.0 &
  91.0 &
  44.0 &
  2.0 &
  54.07 &
  69.49 &
  183.38 &
  251.16 &
  0.12 &
  0.14 &
  0.07 &
  0.07 &
  416.37 &
  456.39 &
  721.84 &
  1538.51 \\ \hline
\textbf{PICO} &
  {\color[HTML]{009901} \textbf{2.98}} &
  {\color[HTML]{009901} \textbf{3.98}} &
  {\color[HTML]{009901} \textbf{4.96}} &
  {\color[HTML]{009901} \textbf{8.0}} &
  {\color[HTML]{CB0000} \textbf{0.0}} &
  {\color[HTML]{009901} \textbf{0.02}} &
  {\color[HTML]{009901} \textbf{0.02}} &
  {\color[HTML]{009901} \textbf{0.03}} &
  {\color[HTML]{CB0000} \textbf{100.0}} &
  {\color[HTML]{009901} \textbf{95.0}} &
  {\color[HTML]{009901} \textbf{57.0}} &
  {\color[HTML]{009901} \textbf{7.0}} &
  {\color[HTML]{009901} \textbf{30.83}} &
  {\color[HTML]{009901} \textbf{49.06}} &
  {\color[HTML]{009901} \textbf{144.67}} &
  {\color[HTML]{009901} \textbf{240.15}} &
  {\color[HTML]{009901} \textbf{0.1}} &
  {\color[HTML]{009901} \textbf{0.08}} &
  {\color[HTML]{009901} \textbf{0.03}} &
  {\color[HTML]{009901} \textbf{0.03}} &
  {\color[HTML]{009901} \textbf{251.39}} &
  {\color[HTML]{009901} \textbf{298.55}} &
  {\color[HTML]{009901} \textbf{526.4}} &
  {\color[HTML]{009901} \textbf{1291.71}} \\ \hline
 &
  \multicolumn{24}{c|}{\textbf{32 Agents (0\%,10\%,20\%,30\% Obstacle Densities)}} \\ \cline{2-25} 
{\textbf{Methods}} &
  \multicolumn{4}{c|}{\textbf{CA $\downarrow$}} &
  \multicolumn{4}{c|}{\textbf{CO $\downarrow$}} &
  \multicolumn{4}{c|}{\textbf{SR $\uparrow$}} &
  \multicolumn{4}{c|}{\textbf{MS $\downarrow$}} &
  \multicolumn{4}{c|}{\textbf{CR $\downarrow$}} &
  \multicolumn{4}{c|}{\textbf{TM $\downarrow$}} \\ \hline
\textbf{ODrM*~\cite{Ferner2013ODrMOM}} &
  {\color[HTML]{CB0000} \textbf{0.0}} &
  {\color[HTML]{CB0000} \textbf{0.0}} &
  {\color[HTML]{CB0000} \textbf{0.0}} &
  {\color[HTML]{CB0000} \textbf{0.0}} &
  {\color[HTML]{CB0000} \textbf{0.0}} &
  {\color[HTML]{CB0000} \textbf{0.0}} &
  {\color[HTML]{CB0000} \textbf{0.0}} &
  {\color[HTML]{CB0000} \textbf{0.0}} &
  {\color[HTML]{CB0000} \textbf{100.0}} &
  {\color[HTML]{CB0000} \textbf{100.0}} &
  {\color[HTML]{CB0000} \textbf{100.0}} &
  {\color[HTML]{CB0000} \textbf{91.0}} &
  {\color[HTML]{CB0000} \textbf{29.09}} &
  {\color[HTML]{CB0000} \textbf{29.67}} &
  {\color[HTML]{CB0000} \textbf{32.88}} &
  {\color[HTML]{CB0000} \textbf{38.93}} &
  {\color[HTML]{CB0000} \textbf{0.0}} &
  {\color[HTML]{CB0000} \textbf{0.0}} &
  {\color[HTML]{CB0000} \textbf{0.0}} &
  {\color[HTML]{CB0000} \textbf{0.0}} &
  {\color[HTML]{CB0000} \textbf{441.34}} &
  {\color[HTML]{CB0000} \textbf{462.07}} &
  {\color[HTML]{CB0000} \textbf{505.13}} &
  {\color[HTML]{CB0000} \textbf{530.07}} \\ \hline
\textbf{PRIMAL~\cite{Sartoretti2019PRIMALPV}} &
  26.23 &
  30.48 &
  47.27 &
  98.33 &
  {\color[HTML]{CB0000} \textbf{0.0}} &
  0.36 &
  1.56 &
  2.07 &
  92.0 &
  72.0 &
  9.0 &
  {\color[HTML]{009901} \textbf{0.0}} &
  53.91 &
  108.02 &
  244.76 &
  {\color[HTML]{009901} \textbf{256.0}} &
  0.47 &
  0.28 &
  0.19 &
  0.38 &
  958.39 &
  1094.29 &
  2226.83 &
  3431.49 \\ \hline
\textbf{DHC~\cite{Ma2021DistributedHM}} &
  27.04 &
  34.79 &
  49.29 &
  95.51 &
  0.02 &
  0.65 &
  4.28 &
  4.67 &
  92.0 &
  62.0 &
  3.0 &
  {\color[HTML]{009901} \textbf{0.0}} &
  48.76 &
  135.78 &
  242.67 &
  {\color[HTML]{009901} \textbf{256.0}} &
  0.52 &
  0.26 &
  0.21 &
  0.37 &
  957.06 &
  1144.55 &
  2009.0 &
  3742.41 \\ \hline
\textbf{PICO-heuristic} &
  27.49 &
  31.01 &
  49.02 &
  93.5 &
  0.01 &
  0.64 &
  3.76 &
  5.18 &
  91.0 &
  63.0 &
  0.0 &
  {\color[HTML]{009901} \textbf{0.0}} &
  49.45 &
  121.05 &
  256.0 &
  {\color[HTML]{009901} \textbf{256.0}} &
  0.56 &
  0.27 &
  0.19 &
  0.36 &
  1144.46 &
  1179.09 &
  2236.79 &
  3723.63 \\ \hline
\textbf{PICO} &
  {\color[HTML]{009901} \textbf{14.8}} &
  {\color[HTML]{009901} \textbf{20.62}} &
  {\color[HTML]{009901} \textbf{36.28}} &
  {\color[HTML]{009901} \textbf{83.38}} &
  {\color[HTML]{CB0000} \textbf{0.0}} &
  {\color[HTML]{009901} \textbf{0.21}} &
  {\color[HTML]{009901} \textbf{1.28}} &
  {\color[HTML]{009901} \textbf{1.63}} &
  {\color[HTML]{CB0000} \textbf{100.0}} &
  {\color[HTML]{009901} \textbf{75.0}} &
  {\color[HTML]{009901} \textbf{19.0}} &
  {\color[HTML]{009901} \textbf{0.0}} &
  {\color[HTML]{009901} \textbf{38.09}} &
  {\color[HTML]{009901} \textbf{96.5}} &
  {\color[HTML]{009901} \textbf{224.54}} &
  {\color[HTML]{009901} \textbf{256.0}} &
  {\color[HTML]{009901} \textbf{0.4}} &
  {\color[HTML]{009901} \textbf{0.22}} &
  {\color[HTML]{009901} \textbf{0.15}} &
  {\color[HTML]{009901} \textbf{0.33}} &
  {\color[HTML]{009901} \textbf{550.87}} &
  {\color[HTML]{009901} \textbf{774.01}} &
  {\color[HTML]{009901} \textbf{1712.7}} &
  {\color[HTML]{009901} \textbf{3175.57}} \\ \hline
 &
  \multicolumn{24}{c|}{\textbf{64 Agents (0\%,10\%,20\%,30\% Obstacle Densities)}} \\ \cline{2-25} 
  {\textbf{Methods}} &
  \multicolumn{4}{c|}{\textbf{CA $\downarrow$}} &
  \multicolumn{4}{c|}{\textbf{CO $\downarrow$}} &
  \multicolumn{4}{c|}{\textbf{SR $\uparrow$}} &
  \multicolumn{4}{c|}{\textbf{MS $\downarrow$}} &
  \multicolumn{4}{c|}{\textbf{CR $\downarrow$}} &
  \multicolumn{4}{c|}{\textbf{TM $\downarrow$}} \\ \hline
\textbf{ODrM*~\cite{Ferner2013ODrMOM}} &
  {\color[HTML]{CB0000} \textbf{0.0}} &
  {\color[HTML]{CB0000} \textbf{0.0}} &
  {\color[HTML]{CB0000} \textbf{0.0}} &
  {\color[HTML]{CB0000} \textbf{0.0}} &
  {\color[HTML]{CB0000} \textbf{0.0}} &
  {\color[HTML]{CB0000} \textbf{0.0}} &
  {\color[HTML]{CB0000} \textbf{0.0}} &
  {\color[HTML]{CB0000} \textbf{0.0}} &
  {\color[HTML]{CB0000} \textbf{100.0}} &
  {\color[HTML]{CB0000} \textbf{96.0}} &
  {\color[HTML]{CB0000} \textbf{79.0}} &
  {\color[HTML]{CB0000} \textbf{17.0}} &
  {\color[HTML]{CB0000} \textbf{31.82}} &
  {\color[HTML]{CB0000} \textbf{31.87}} &
  {\color[HTML]{CB0000} \textbf{29.82}} &
  {\color[HTML]{CB0000} \textbf{7.53}} &
  {\color[HTML]{CB0000} \textbf{0.0}} &
  {\color[HTML]{CB0000} \textbf{0.0}} &
  {\color[HTML]{CB0000} \textbf{0.0}} &
  {\color[HTML]{CB0000} \textbf{0.0}} &
  {\color[HTML]{CB0000} \textbf{916.53}} &
  {\color[HTML]{CB0000} \textbf{936.43}} &
  {\color[HTML]{CB0000} \textbf{844.11}} &
  {\color[HTML]{CB0000} \textbf{202.05}} \\ \hline
\textbf{PRIMAL~\cite{Sartoretti2019PRIMALPV}} &
  115.79 &
  171.31 &
  341.95 &
  634.7 &
  0.11 &
  2.27 &
  8.04 &
  26.08 &
  75.0 &
  7.0 &
  {\color[HTML]{009901} \textbf{0.0}} &
  {\color[HTML]{009901} \textbf{0.0}} &
  111.27 &
  241.73 &
  {\color[HTML]{009901} \textbf{256.0}} &
  {\color[HTML]{009901} \textbf{256.0}} &
  1.04 &
  0.71 &
  1.34 &
  2.48 &
  2418.58 &
  3679.7 &
  6611.44 &
  9156.88 \\ \hline
\textbf{DHC~\cite{Ma2021DistributedHM}} &
  106.78 &
  146.92 &
  312.92 &
  622.58 &
  0.25 &
  12.75 &
  38.37 &
  145.89 &
  72.0 &
  0.0 &
  {\color[HTML]{009901} \textbf{0.0}} &
  {\color[HTML]{009901} \textbf{0.0}} &
  108.76 &
  256.0 &
  {\color[HTML]{009901} \textbf{256.0}} &
  {\color[HTML]{009901} \textbf{256.0}} &
  0.98 &
  0.57 &
  1.22 &
  2.43 &
  2121.45 &
  3261.68 &
  6091.9 &
  8407.14 \\ \hline
\textbf{PICO-heuristic} &
  112.99 &
  142.03 &
  303.23 &
  621.68 &
  0.25 &
  12.75 &
  38.37 &
  145.89 &
  70.0 &
  3.0 &
  {\color[HTML]{009901} \textbf{0.0}} &
  {\color[HTML]{009901} \textbf{0.0}} &
  108.76 &
  243.36 &
  {\color[HTML]{009901} \textbf{256.0}} &
  {\color[HTML]{009901} \textbf{256.0}} &
  1.03 &
  0.58 &
  1.18 &
  2.42 &
  2201.45 &
  3246.68 &
  6162.9 &
  8338.14 \\ \hline
\textbf{PICO} &
  {\color[HTML]{009901} \textbf{90.95}} &
  {\color[HTML]{009901} \textbf{128.4}} &
  {\color[HTML]{009901} \textbf{279.61}} &
  {\color[HTML]{009901} \textbf{591.14}} &
  {\color[HTML]{009901} \textbf{0.36}} &
  {\color[HTML]{009901} \textbf{8.76}} &
  {\color[HTML]{009901} \textbf{38.36}} &
  {\color[HTML]{009901} \textbf{130.27}} &
  {\color[HTML]{009901} \textbf{83.0}} &
  {\color[HTML]{009901} \textbf{13.0}} &
  {\color[HTML]{009901} \textbf{0.0}} &
  {\color[HTML]{009901} \textbf{0.0}} &
  {\color[HTML]{009901} \textbf{94.36}} &
  {\color[HTML]{009901} \textbf{224.88}} &
  {\color[HTML]{009901} \textbf{256.0}} &
  {\color[HTML]{009901} \textbf{256.0}} &
  {\color[HTML]{009901} \textbf{0.96}} &
  {\color[HTML]{009901} \textbf{0.55}} &
  {\color[HTML]{009901} \textbf{1.09}} &
  {\color[HTML]{009901} \textbf{2.31}} &
  {\color[HTML]{009901} \textbf{1472.92}} &
  {\color[HTML]{009901} \textbf{2621.25}} &
  {\color[HTML]{009901} \textbf{5342.04}} &
  {\color[HTML]{009901} \textbf{7713.69}} \\ \hline
\end{tabular}%
}
\end{table*}

\begin{figure*}[htb!]
    \centering
    \includegraphics[width=\textwidth]{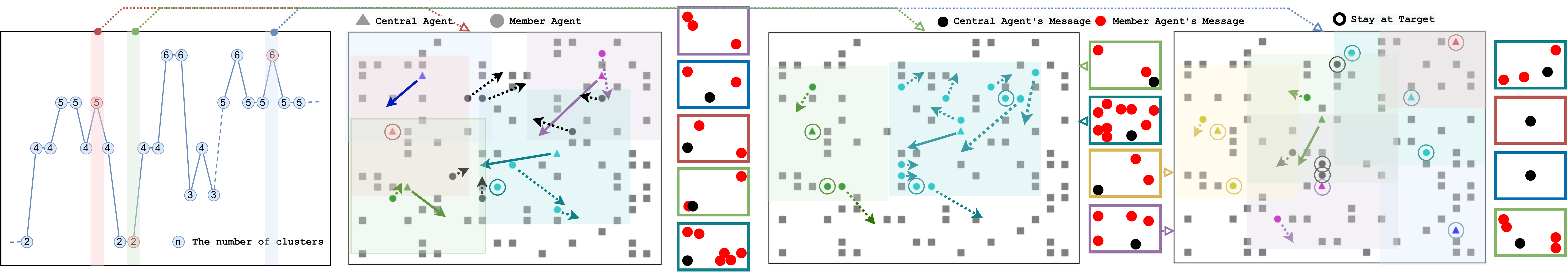}
    \caption{The cluster's composition changes over time in a $20$-sized, $16$ agents, and $0.3$ obstacle density environment. The arrow indicates the target's direction, and its length is proportional to the distance. The message delivered by the agent is also shown in the figure. We use t-SNE~\cite{Maaten2014AcceleratingTU} to reduce the message to $2$ dimensions and regularize it to the range of $0$ to $1$.}
    \label{fig:clusters}
    \vspace{-15pt}
\end{figure*}

\noindent{\em{Results Analysis.}}
Table~\ref{tab:main-rst} shows six measurement comparisons of PICO and baselines in a square gridworld with a size of $20$, different obstacle densities ($0$, $0.1$, $0.2$, $0.3$), and the number of agents ($8$, $16$, $32$, $64$).
The collision rates between PICO agents in various scenarios are significantly lower than the baselines.
It can directly found from these experimental results that PICO has better generalization.
PICO is trained only in the $8$-agents scenario, but PICO can still maintain good performance in a new environment that is expanded to $\sim10$ times the number of agents.

\noindent{\em{Ablation Studies}}.
To further analyze the importance of priority and the effectiveness of implicit priority learning, we conduct ablation studies to verify the impact of different priorities on the algorithm's performance.
Specifically, we have designed a greedy priority allocation strategy.
At each timestep $t$, we extract the current Manhattan distance to the goal 
and denote it as $mh_t^i$.
At the same time, for each pair of agents $i,j$ with the same goal distance $mh_t^i=mh_t^j$, if the initial goal distance $mh_0^j$ of the agent $j$ is farther from than $mh_0^i$, then $j$ has a higher priority. 
Therefore, for any agent $i$, the priority is calculated by $p_t^i = - mh_t^i / mh_0^i$.
If the two agents have the same highest priority, one of them will be randomly selected to be the high-level agent.
This ablation version of PICO that uses the above greedy priority strategy is denoted as PICO-heuristic.
The comparison results are also shown in Table~\ref{tab:main-rst}.
The experimental results well verify the importance of priority in communication learning and the effectiveness of implicit priority learning.

\noindent{\em{Diving Into the Path Planning}}.
We visually analyze the 
cluster composition of PICO during path planning.
Specifically, we count the changes of the cluster numbers formed by CBRP based on implicit priority during PICO's path planning, while the results are shown in Figure~\ref{fig:clusters}.
In the beginning, agents are generally uniformly distributed in the environment, 
so there will be more clusters formed;
As the path planning progresses, the agents begin to intersect each other more,
and the number of clusters will be reduced accordingly;
Finally, the agents are closer to the goals, 
which makes the agents farther apart again.
We can find that the farther away from the target, the more likely it is to become a central agent.
This figure also shows more diverse central agent selection strategies.

\section{CONCLUSION}\label{sec:con}

In this paper, we propose the prioritized communication learning method to incorporate the implicit planning priority into the communication topology within the communication-based MARL framework.
PICO makes a trade-off between decentralized path planning but (nearly) centralized collision avoiding planning, and achieves better collision avoidance and real-time performance simultaneously. 
Experiments show that PICO performs significantly better in large-scale pathfinding tasks in both success rates and collision rates than state-of-the-art learning-based planners.

\addtolength{\textheight}{-3.6cm}   





\section*{ACKNOWLEDGMENT}

This work was supported in part by the National Key Research and Development Program of China~(No. 2020AAA0107400), NSFC (No. 12071145), STCSM (No. 19ZR141420), the Open Research Projects of Zhejiang Lab (NO.2021KE0AB03), and the Fundamental Research Funds for the Central Universities.





\bibliographystyle{IEEEtran}
\bibliography{IEEEabrv,main}   

\begin{thebibliography}{10}
\providecommand{\url}[1]{#1}
\csname url@rmstyle\endcsname
\providecommand{\newblock}{\relax}
\providecommand{\bibinfo}[2]{#2}
\providecommand\BIBentrySTDinterwordspacing{\spaceskip=0pt\relax}
\providecommand\BIBentryALTinterwordstretchfactor{4}
\providecommand\BIBentryALTinterwordspacing{\spaceskip=\fontdimen2\font plus
\BIBentryALTinterwordstretchfactor\fontdimen3\font minus
  \fontdimen4\font\relax}
\providecommand\BIBforeignlanguage[2]{{%
\expandafter\ifx\csname l@#1\endcsname\relax
\typeout{** WARNING: IEEEtran.bst: No hyphenation pattern has been}%
\typeout{** loaded for the language `#1'. Using the pattern for}%
\typeout{** the default language instead.}%
\else
\language=\csname l@#1\endcsname
\fi
#2}}

\bibitem{Rubenstein2014ProgrammableSI}
M.~Rubenstein, A.~Cornejo, and R.~Nagpal, ``Programmable self-assembly in a
  thousand-robot swarm,'' \emph{Science}, vol. 345, pp. 795--799, 2014.

\bibitem{Howard2006ExperimentsWA}
A.~Howard, L.~Parker, and G.~Sukhatme, ``Experiments with a large heterogeneous
  mobile robot team: Exploration, mapping, deployment and detection,''
  \emph{The International Journal of Robotics Research}, vol.~25, pp. 431--447,
  2006.

\bibitem{Silver2005CooperativeP}
D.~Silver, ``Cooperative pathfinding,'' in \emph{AIIDE}, 2005.

\bibitem{Berg2009ReciprocalNC}
J.~V.~D. Berg, S.~Guy, M.~Lin, and D.~Manocha, ``Reciprocal n-body collision
  avoidance,'' in \emph{ISRR}, 2009.

\bibitem{Sharon2012ConflictbasedSF}
G.~Sharon, R.~Stern, A.~Felner, and N.~R. Sturtevant, ``Conflict-based search
  for optimal multi-agent pathfinding,'' \emph{Artificial Intelligence}, vol.
  219, pp. 40--66, 2012.

\bibitem{Felner2018AddingHT}
A.~Felner, J.~Li, E.~Boyarski, H.~Ma, L.~Cohen, T.~K.~S. Kumar, and S.~Koenig,
  ``Adding heuristics to conflict-based search for multi-agent path finding,''
  in \emph{ICAPS}, 2018.

\bibitem{Sartoretti2019PRIMALPV}
G.~Sartoretti, J.~Kerr, Y.~Shi, G.~Wagner, T.~K.~S. Kumar, S.~Koenig, and
  H.~Choset, ``Primal: Pathfinding via reinforcement and imitation multi-agent
  learning,'' \emph{IEEE Robotics and Automation Letters}, vol.~4, pp.
  2378--2385, 2019.

\bibitem{Zhang2020LearningTC}
Y.~Zhang, Y.~Qian, Y.~Yao, H.~Hu, and Y.~Xu, ``Learning to cooperate:
  Application of deep reinforcement learning for online {AGV} path finding,''
  in \emph{AAMAS}, 2020.

\bibitem{Damani2021PRIMAL_2PV}
M.~Damani, Z.~Luo, E.~Wenzel, and G.~Sartoretti, ``Primal$_2$: Pathfinding via
  reinforcement and imitation multi-agent learning - lifelong,'' \emph{IEEE
  Robotics and Automation Letters}, vol.~6, pp. 2666--2673, 2021.

\bibitem{Stern2019MultiAgentPD}
R.~Stern, N.~R. Sturtevant, A.~Felner, S.~Koenig, H.~Ma, T.~T. Walker, J.~Li,
  D.~Atzmon, L.~Cohen, T.~K.~S. Kumar, E.~Boyarski, and R.~Bart{\'a}k,
  ``Multi-agent pathfinding: Definitions, variants, and benchmarks,'' in
  \emph{SOCS}, 2019.

\bibitem{Ferner2013ODrMOM}
C.~Ferner, G.~Wagner, and H.~Choset, ``{ODrM}* optimal multirobot path planning
  in low dimensional search spaces,'' in \emph{ICRA}, 2013.

\bibitem{Sartoretti2018DistributedRL}
G.~Sartoretti, Y.~Wu, W.~Paivine, T.~K.~S. Kumar, S.~Koenig, and H.~Choset,
  ``Distributed reinforcement learning for multi-robot decentralized collective
  construction,'' in \emph{DARS}, 2018.

\bibitem{Lowe2019OnTP}
R.~Lowe, J.~Foerster, Y.~Boureau, J.~Pineau, and Y.~Dauphin, ``On the pitfalls
  of measuring emergent communication,'' in \emph{AAMAS}, 2019.

\bibitem{Ma2019SearchingWC}
H.~Ma, D.~Harabor, P.~Stuckey, J.~Li, and S.~Koenig, ``Searching with
  consistent prioritization for multi-agent path finding,'' in \emph{SOCS},
  2019.

\bibitem{Sturtevant2006ImprovingCP}
N.~R. Sturtevant and M.~Buro, ``Improving collaborative pathfinding using map
  abstraction,'' in \emph{AIIDE}, 2006.

\bibitem{Velagapudi2010DecentralizedPP}
P.~Velagapudi, K.~Sycara, and P.~Scerri, ``Decentralized prioritized planning
  in large multirobot teams,'' in \emph{IROS}, 2010.

\bibitem{Wang2011MAPPAS}
K.~Wang and A.~Botea, ``{MAPP}: a scalable multi-agent path planning algorithm
  with tractability and completeness guarantees,'' \emph{Journal of Artificial
  Intelligence Research}, vol.~42, pp. 55--90, 2011.

\bibitem{Zhiyao2020DeepRL}
L.~Zhiyao and G.~Sartoretti, ``Deep reinforcement learning based multi-agent
  pathfinding,'' \emph{Technical Report.}, 2020.

\bibitem{freed2020simultaneous}
B.~Freed, G.~Sartoretti, and H.~Choset, ``Simultaneous policy and discrete
  communication learning for multi-agent cooperation,'' \emph{IEEE Robotics and
  Automation Letters}, vol.~5, no.~2, pp. 2498--2505, 2020.

\bibitem{Ma2021DistributedHM}
Z.~Ma, Y.~Luo, and H.~Ma, ``Distributed heuristic multi-agent path finding with
  communication,'' in \emph{ICRA}, 2021.

\bibitem{Erdmann1987MultipleMoving}
E.~Michael and T.~Lozano-Perez, ``On multiple moving objects,''
  \emph{Algorithmica}, vol.~2, no.~1, pp. 477--521, 1987.

\bibitem{Warren1990MultipleRP}
C.~W. Warren, ``Multiple robot path coordination using artificial potential
  fields,'' in \emph{ICRA}, 1990.

\bibitem{Bennewitz2002FindingAO}
M.~Bennewitz, W.~Burgard, and S.~Thrun, ``Finding and optimizing solvable
  priority schemes for decoupled path planning techniques for teams of mobile
  robots,'' \emph{Robotics and Autonomous Systems}, vol.~41, pp. 89--99, 2002.

\bibitem{Berg2005PrioritizedMP}
J.~V.~D. Berg and M.~Overmars, ``Prioritized motion planning for multiple
  robots,'' in \emph{IROS}, 2005.

\bibitem{Buckley1989FastMP}
S.~Buckley, ``Fast motion planning for multiple moving robots,'' in
  \emph{ICRA}, 1989.

\bibitem{Ferrari1998MultirobotMC}
C.~Ferrari, E.~Pagello, J.~Ota, and T.~Arai, ``Multirobot motion coordination
  in space and time,'' \emph{Robotics and Autonomous Systems}, vol.~25, pp.
  219--229, 1998.

\bibitem{ODonnell1989DeadlockfreeAC}
P.~A. O'Donnell and T.~Lozano-Perez, ``Deadlock-free and collision-free
  coordination of two robot manipulators,'' in \emph{ICRA}, 1989.

\bibitem{Azarm1997ConflictfreeMO}
K.~Azarm and G.~Schmidt, ``Conflict-free motion of multiple mobile robots based
  on decentralized motion planning and negotiation,'' in \emph{ICRA}, 1997.

\bibitem{hansen2004dynamic}
E.~A. Hansen, D.~S. Bernstein, and S.~Zilberstein, ``Dynamic programming for
  partially observable stochastic games,'' in \emph{AAAI}, 2004.

\bibitem{Levine2014GuidedPS}
S.~Levine and V.~Koltun, ``Learning complex neural network policies with
  trajectory optimization,'' in \emph{ICML}, 2014.

\bibitem{Levine2014LearningNN}
S.~Levine and P.~Abbeel, ``Learning neural network policies with guided policy
  search under unknown dynamics,'' in \emph{NeurIPS}, 2014.

\bibitem{Mnih2015HumanlevelCT}
V.~Mnih, K.~Kavukcuoglu, D.~Silver, A.~A. Rusu, J.~Veness, M.~G. Bellemare,
  A.~Graves, M.~A. Riedmiller, A.~Fidjeland, G.~Ostrovski, S.~Petersen,
  C.~Beattie, A.~Sadik, I.~Antonoglou, H.~King, D.~Kumaran, D.~Wierstra,
  S.~Legg, and D.~Hassabis, ``Human-level control through deep reinforcement
  learning,'' \emph{Nature}, vol. 518, pp. 529--533, 2015.

\bibitem{Grill2020BootstrapYO}
J.-B. Grill, F.~Strub, F.~Altch'e, C.~Tallec, P.~H. Richemond, E.~Buchatskaya,
  C.~Doersch, B.~A. Pires, Z.~Guo, M.~G. Azar, B.~Piot, K.~Kavukcuoglu,
  R.~Munos, and M.~Valko, ``Bootstrap your own latent: A new approach to
  self-supervised learning,'' in \emph{NeurIPS}, 2020.

\bibitem{shoham2008multiagent}
Y.~Shoham and K.~Leyton-Brown, \emph{Multiagent systems: Algorithmic,
  game-theoretic, and logical foundations}.\hskip 1em plus 0.5em minus
  0.4em\relax Cambridge University Press, 2008.

\bibitem{Peysakhovich2018ConsequentialistCC}
A.~Peysakhovich and A.~Lerer, ``Consequentialist conditional cooperation in
  social dilemmas with imperfect information,'' in \emph{AAAI Workshops}, 2018.

\bibitem{Peysakhovich2018ProsocialLA}
------, ``Prosocial learning agents solve generalized stag hunts better than
  selfish ones,'' in \emph{AAMAS}, 2018.

\bibitem{Hughes2018InequityAI}
E.~Hughes, J.~Z. Leibo, M.~Phillips, K.~Tuyls, E.~A. Du{\'e}{\~n}ez-Guzm{\'a}n,
  A.~Casta{\~n}eda, I.~Dunning, T.~Zhu, K.~R. McKee, R.~Koster, H.~Roff, and
  T.~Graepel, ``Inequity aversion improves cooperation in intertemporal social
  dilemmas,'' in \emph{NeurIPS}, 2018.

\bibitem{Jaques2019SocialIA}
N.~Jaques, A.~Lazaridou, E.~Hughes, Çaglar G{\"u}lçehre, P.~A. Ortega,
  D.~Strouse, J.~Z. Leibo, and N.~D. Freitas, ``Social influence as intrinsic
  motivation for multi-agent deep reinforcement learning,'' in \emph{ICML},
  2019.

\bibitem{cbrp}
M.~Rezaee and M.~Yaghmaee, ``Cluster based routing protocol for mobile ad hoc
  networks,'' \emph{INFOCOM}, vol.~8, no.~1, pp. 30--36, 2009.

\bibitem{Mnih2016AsynchronousMF}
V.~Mnih, A.~P. Badia, M.~Mirza, A.~Graves, T.~Lillicrap, T.~Harley, D.~Silver,
  and K.~Kavukcuoglu, ``Asynchronous methods for deep reinforcement learning,''
  in \emph{ICML}, 2016.

\bibitem{Vinyals2017StarCraftIA}
O.~Vinyals, T.~Ewalds, S.~Bartunov, P.~Georgiev, A.~Vezhnevets, M.~Yeo,
  A.~Makhzani, H.~K{\"u}ttler, J.~Agapiou, J.~Schrittwieser, J.~Quan,
  S.~Gaffney, S.~Petersen, K.~Simonyan, T.~Schaul, H.~V. Hasselt, D.~Silver,
  T.~Lillicrap, K.~Calderone, P.~Keet, A.~Brunasso, D.~Lawrence, A.~Ekermo,
  J.~Repp, and R.~Tsing, ``Starcraft {II}: A new challenge for reinforcement
  learning,'' \emph{ArXiv}, vol. abs/1708.04782, 2017.

\bibitem{Haarnoja2018SoftAO}
T.~Haarnoja, A.~Zhou, P.~Abbeel, and S.~Levine, ``Soft actor-critic: Off-policy
  maximum entropy deep reinforcement learning with a stochastic actor,'' in
  \emph{ICML}, 2018.

\bibitem{Nguyen2017GeneralizedTA}
V.~Nguyen, P.~Obermeier, T.~C. Son, T.~Schaub, and W.~Yeoh, ``Generalized
  target assignment and path finding using answer set programming,'' in
  \emph{IJCAI}, 2017.

\bibitem{Wagner2015SubdimensionalEF}
G.~Wagner and H.~Choset, ``Subdimensional expansion for multirobot path
  planning,'' \emph{Artificial Intelligence}, vol. 219, pp. 1--24, 2015.

\bibitem{Maaten2014AcceleratingTU}
L.~V.~D. Maaten, ``Accelerating t-sne using tree-based algorithms,'' \emph{J.
  Mach. Learn. Res.}, vol.~15, pp. 3221--3245, 2014.

\end{thebibliography}

\end{document}